\documentclass{llncs}
\usepackage{float}
\usepackage{tabularx,ragged2e}
\usepackage{amsmath}
\usepackage{booktabs}
\usepackage{multirow}
\usepackage{subfigure}
\usepackage{lscape}
\usepackage[table,xcdraw]{xcolor}
\usepackage{makeidx}  
 \usepackage{multirow}
 \usepackage[table,xcdraw]{xcolor}
\usepackage{amssymb}
\usepackage{array}
\newcolumntype{P}[1]{>{\centering\arraybackslash}p{#1}}
\usepackage{graphicx}
\usepackage{tabu}
\usepackage[flushleft]{threeparttable}
\usepackage{url}
\usepackage{algorithm}
\usepackage[noend]{algpseudocode}
\makeindex
\begin{document}

\title{Quantifying the Effect of Image Similarity on Diabetic Foot Ulcer Classification}

\author{Imran Chowdhury Dipto\inst{1}\orcidID{0000-0001-9183-2651} 
\and Bill Cassidy \inst{1}\thanks{equal contribution}\orcidID{0000-0003-3741-8120}  
\and Connah Kendrick \inst{1}\orcidID{0000-0002-3623-6598}  
\and Neil D. Reeves \inst{3}\orcidID{0000-0001-9213-4580} 
\and Joseph M. Pappachan \inst{2}\orcidID{0000-0003-0886-5255}
\and Vishnu Chandrabalan \inst{2}\orcidID{0000-0002-2687-1096}  
\and Moi Hoon Yap \inst{1}\orcidID{0000-0001-7681-4287}}
%
\authorrunning{I. Chowdhury et al.}
%

\institute{
	Centre for Advanced Computational Science, 
	Department of Computing and Mathematics,
	Manchester Metropolitan University, 
	Manchester M1 5GD, United Kingdom \and
	Lancashire Teaching Hospitals NHS Trust, 
	Preston, PR2 9HT, United Kingdom \and
	Musculoskeletal Science and Sports Medicine Research Centre, 
	Manchester Metropolitan University, 
	Manchester M1 5GD, United Kingdom  \\
	\email{M.Yap@mmu.ac.uk}
}

\maketitle


\begin{abstract}
	This research conducts an investigation on the effect of visually similar images within a publicly available diabetic foot ulcer dataset when training deep learning classification networks.
	The presence of binary-identical duplicate images in datasets used to train deep learning algorithms is a well known issue that can introduce unwanted bias which can degrade network performance. However, the effect of visually similar non-identical images is an under-researched topic, and has so far not been investigated in any diabetic foot ulcer studies. 
	We use an open-source fuzzy algorithm to identify groups of increasingly similar images in the Diabetic Foot Ulcers Challenge 2021 (DFUC2021) training dataset. Based on each similarity threshold, we create new training sets that we use to train a range of deep learning multi-class classifiers. We then evaluate the performance of the best performing model on the DFUC2021 test set.
	Our findings show that the model trained on the training set with the 80\% similarity threshold images removed achieved the best performance using the InceptionResNetV2 network. This model showed improvements in F1-score, precision, and recall of 0.023, 0.029, and 0.013, respectively.
	These results indicate that highly similar images can contribute towards the presence of performance degrading bias within the Diabetic Foot Ulcers Challenge 2021 dataset, and that the removal of images that are 80\% similar from the training set can help to boost classification performance.
\end{abstract}

\section{Introduction}
Since the publication of the DFUC2021 Proceedings, there has been no substantial progress made on the DFU multi-class classification task. This paper studies one of possible cause, i.e., the effect of visually similar non-identical images within DFUC2021 dataset. Image duplication (the presence of binary identical images) is generally acknowledged as a factor in reducing model performance when training deep learning models due to the performance degrading bias that over-represented features may introduce into the trained model. However, the effect of visually similar non-identical images which may result in an over-representation of certain features present in deep learning datasets is an under-researched topic. An overabundance of certain features in a dataset may cause undesirable performance degrading bias in any models trained using them. In this paper we conduct an analysis of the effect of images that are visually similar but not binary identical on a publicly available diabetic foot ulcer dataset using an open-source fuzzy matching algorithm. We train a large range of multi-class deep learning classification models on the Diabetic Foot Ulcer Challenge 2021 dataset (DFUC2021) \cite{yap2021analysis}, and for the first time, quantify the effect of image similarity on network accuracy.

We found no studies that observed and quantified the effect of feature over-representation or the effect of image similarity in DFU research. The effect of binary duplicate images has been observed in other domains \cite{cassidy2022isic} but the topic remains an under-researched problem generally. A common theme with many previous studies is limited dataset size. A small dataset may hinder the ability of models to generalise to a wider range of examples in real-world settings. Conversely, a large dataset might introduce performance degrading feature-bias if the data has been collected from a small number of subjects. To address this, we conduct experiments to analyse the effect of image similarity on the DFUC2021 dataset.

\section{Related Work} \label{related_work}

Previous research on DFU has involved localisation \cite{goyal2018robust,cassidy2020dfuc,yap2021detection,reeves2021diabetes,cassidy2021cloudbased,joseph2022future}, binary \cite{goyal2020recognition} and multi-class classification \cite{cassidy2021eval,yap2022overview}, and segmentation \cite{goyal2017fully,kendrick2022segmentation}. Goyal et al. \cite{goyal2018dfunet} proposed a deep learning architecture for DFU classification. Their work was notable for achieving high scores in sensitivity and accuracy using a small DFU dataset ($<2000$ images for training and testing).

More recently, Al-Garaawi et al. \cite{algaraawi2022binary} conducted a series of binary classification experiments using DFU patches. This work used mapped local binary pattern coded images with RGB images as inputs to the CNN to increase binary classifier performance. However, a limitation of this work is the use of small datasets.

In last year's Diabetic Foot Ulcers Grand Challenge, Yap et al. \cite{yap2021analysis} conducted multi-class classification experiments using the DFUC2021 dataset. This work highlighted the challenging nature of multi-class classification in this domain due to intra-class similarities.


\section{Methodology} \label{method}

\begin{figure}[!h]
	\centering
	\includegraphics[width=5.8cm,height=5.8cm]{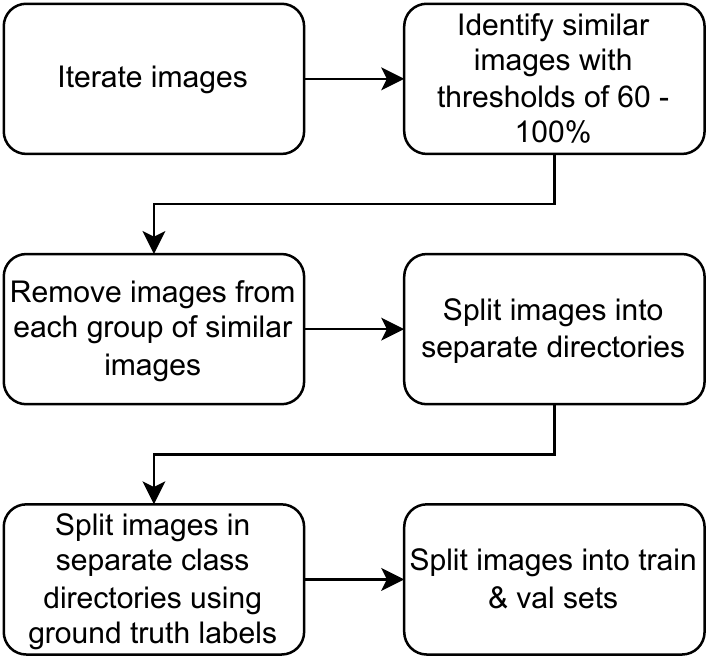}
	\caption{Overview of the process used in the identification and removal of similar images at each similarity threshold.}
	\label{fig:process}
\end{figure}

To observe the effect of similar images in the DFUC2021 dataset on multi-class classification, we devised a strategy of gradually removing successive groups of similar images from the training set. Each group of similar images were identified using the dupeGuru \cite{dupeGuru} Windows application. This open-source application implements a fuzzy search algorithm capable of identifying visually similar images. Results can be filtered by percentage similarity within the application. Using this feature, we were able to identify groups of similar images within the DFUC2021 training set. For each similarity threshold, we train a set of multi-class classifiers capable of classifying the following five classes: (1) control, (2) infection, (3) ischemia, (4) infection and ischemia, and (5) none. Figure \ref{fig:process} shows an overview of the entire process used to create the new training sets used in our experiments.

\subsection{Dataset Description}
For our experiments, we use the publicly available DFUC2021 dataset, introduced by Yap et al. \cite{yap2021analysis}. This dataset is the largest publicly available DFU dataset with wound pathology class labels. The dataset comprises a total of 15,683 images, sized at $224 \times 224$ pixels, with 5955 images for the training set (2555 infection only, 227 ischaemia only, 621 both infection and ischaemia, and 2552 without ischaemia and infection), 3994 unlabeled images, and 5734 images for the testing set. All wounds are cropped from larger images so that only the wound is present in each image. The DFUC2021 dataset is highly heterogeneous due to the nature of the variety of capture devices and variable settings used during photographic acquisition. The ground truth labels were provided by expert clinicians at Lancashire Teaching Hospitals, UK. This dataset was obtained with ethical approval from the UK National Health Service Research Ethics Committee (reference number: 15/NW/0539). 


\subsection{Fuzzy Algorithm}
The fuzzy algorithm used by the dupeGuru application reads each image in RGB bitmap mode which is then split into blocks. Next, the analysis phase uses a $15\times15$ pixel grid to average the colour of each grid tile, the results of which are stored in a cache database. Each grid tile, representing an average colour, is then compared to its corresponding grid on the other image being compared to, and a sum of the difference between R, G and B on each side is computed. The RGB sums are then added together to obtain a final result. If the score is smaller or equal to the user-specified threshold, then a match is found. If a threshold of 100 is set by the user then the algorithm adds an extra constraint indicating that images should contain identical binary data.

\begin{table}[!htbp]
	\centering
	\caption{Summary of the number of similar images found by the dupeGuru fuzzy algorithm in the train, test, and train \& test sets combined at each user-defined similarity threshold.}
	\scalebox{1.00}{
		\begin{tabular}{|l|l|l|l|}
			\hline
			Threshold (\%) & \multicolumn{3}{l|}{Similar Images}  \\
			\cline{2-4}
			& Train Set & Test Set & {Train \& Test Set} \\
			\hline
			\hline
			100 & 0  & 0 & 0 \\ 
			\hline
			95  & 1  & 0 & 3 \\
			\hline
			90  & 19 & 23 & 48 \\
			\hline
			85 & 106 & 125 & 268 \\
			\hline
			80 & 317 & 345 & 719 \\
			\hline
			75 & 590 & 621 & 1278 \\
			\hline
			70 & 1013 & 979 & 2082 \\
			\hline
			65 & 1509 & 1367 & 2976 \\
			\hline
			60 & 2066 & 1683 & 3906 \\
			\hline
		\end{tabular}
	}
	\label{tab:sim}
\end{table}

\subsection{Identification of Similar Images}
To identify the similar images in the DFUC2021 dataset we have used the hardness level (similarity threshold) in the dupeGuru application with the values of 60\%, 65\%, 70\%, 75\%, 80\%, 85\%, 90\%, 95\%, and 100\%. A similarity threshold of 80\%, for example, indicates that the application will find images that have 80\% similarity. We ran the fuzzy algorithm on the training, test, and the training and test sets combined to find similar images that exist exclusively within the training set, exclusively within the test set, and within both train and test sets combined. 
Table \ref{tab:sim} shows a summary of the number of similar images detected by the dupeGuru fuzzy algorithm on the different thresholds in the training set, the test set, and the training and test sets combined.

\subsection{Removal of Similar Images as Determined by Similarity Thresholds}
Images of each of the classes present in the dataset were separated into different directories - one directory per class. On each of these directories the dupeGuru fuzzy algorithm was run using similarity thresholds of 60\% to 100\%.
Next, the filenames for the images in each similarity threshold were saved into CSV files containing Group ID and Image filenames. Group ID refers to the grouping of similar images returned by the fuzzy algorithm, where each group of similar images is assigned a unique sequential identifier. The CSV files output by dupeGuru were then merged with the file containing the ground truth labels of the training set based on the image filenames. For each group of similar images, all but the first image in each group was removed. This meant that a single example from each similarity group was kept for inclusion in each similarity threshold training set. 

\begin{table}[h]
	\caption{Summary of the number of similar images removed at each similarity threshold and the remaining images that are used for the new training sets.}
	\begin{center}
		\begin{tabular}{|l|l|l|} 
			\hline
			Threshold (\%) & Similar Images Removed & Remaining Images \\ [0.5ex] 
			\hline\hline
			95 & 1 & 9948 \\
			\hline
			90 & 19 & 9930 \\
			\hline
			85 & 106 & 9843 \\
			\hline
			80 & 317 & 9632 \\
			\hline
			75 & 590 & 9359 \\
			\hline
			70 & 1013 & 8936 \\
			\hline
			65 & 1508 & 8441 \\
			\hline
			60 & 2068 & 7881 \\
			\hline
		\end{tabular}
	\end{center}
	
	\label{tab:curated_sets}
\end{table}

By comparing the filenames with the ground truth labels, the images in the curated datasets were copied into new directories which formed the new training sets. To check the validity of the results from this process, a Python routine was created which compared the CSV files against the ground truth labels to ensure that the correct images had been copied and that non of the additional images from each similarity group were present in the new training sets. An additional manual spot-check was completed on a random sample of images in the new training sets to ensure that the images had been correctly separated. Table \ref{tab:curated_sets} shows a summary of the number of images removed at each similarity threshold together with the total remaining images used to form the new training sets. 

\section{Image Similarity Analysis}
In this section we analyse a selection of images from each of the similarity thresholds returned by the dupeGuru fuzzy algorithm prior to training the multi-class classification models. Note that not all similarity searches returned results.

\subsection{Train Set Image Similarity}
\begin{figure}[]
	\centering
	\begin{tabular}{cc}
		\includegraphics[width=5cm,height=5cm]{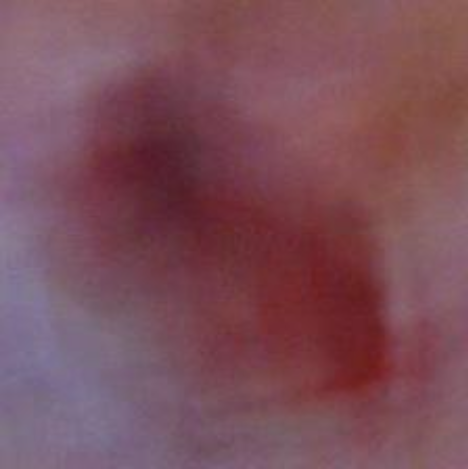} & \includegraphics[width=5cm,height=5cm]{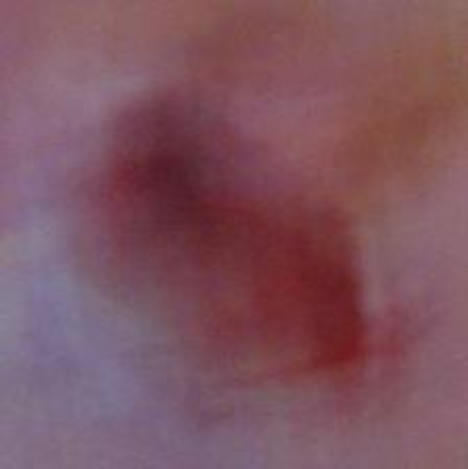} \\
		(a) & (b) \\
		\includegraphics[width=5cm,height=5cm]{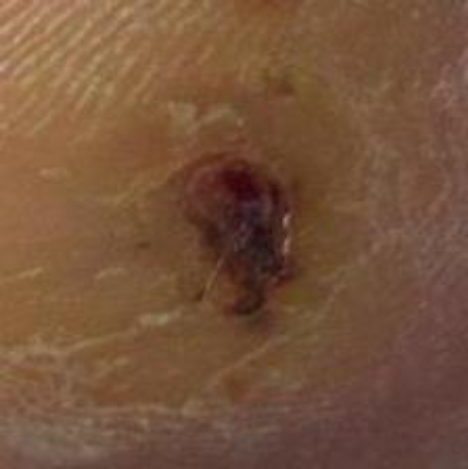} & \includegraphics[width=5cm,height=5cm]{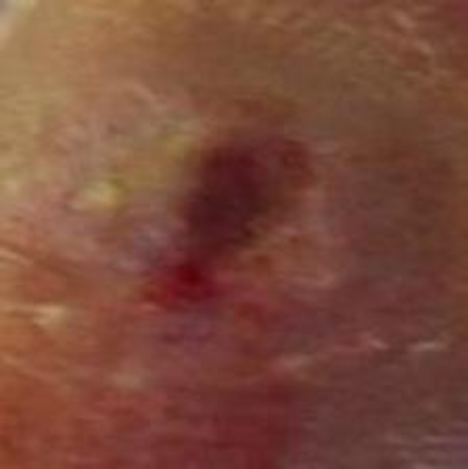} \\
		(c) & (d) \\
		\includegraphics[width=5cm,height=5cm]{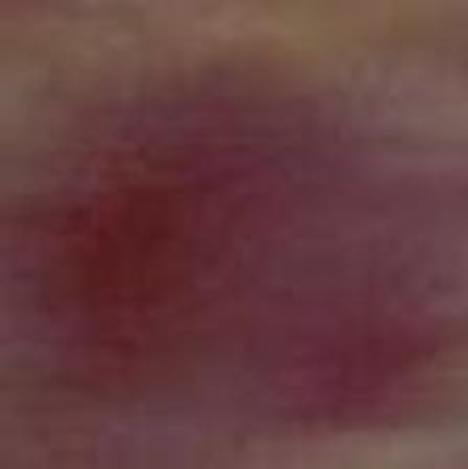} & \includegraphics[width=5cm,height=5cm]{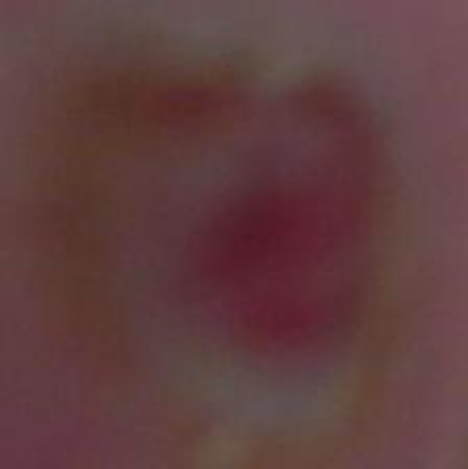} \\
		(e) & (f)
	\end{tabular}
	\caption[]{Illustration of two training set images identified by the dupeGuru fuzzy algorithm with a similarity threshold of 75\% (a \& b), two training set images from the `none' class found in the 65\% similarity threshold (c \& d), and two training set images found in the 60\% similarity threshold - (e) is from the 'none' class, (f) is from the 'unlabelled' class.}
	\label{fig:similar_train_60_to_75}
\end{figure}

Figure \ref{fig:similar_train_60_to_75} (a \& b) shows two images from the training set in the 75\% similarity threshold. These images are of the same wound at different levels of magnification, with the example shown in (a) being at a higher level of magnification. Figure \ref{fig:similar_train_60_to_75} (c \& d) shows two training set images in the 65\% similarity threshold. These two images represent two distinctly different DFU wounds with noticeably different features. Figure \ref{fig:similar_train_60_to_75} (e \& f) shows a further two training set images which were identified in the 60\% similarity threshold. As with the previous examples, these images represent two different wounds.

\subsection{Test Set Image Similarity}
\begin{figure}[!h]
	\centering
	\begin{tabular}{cc}
		\includegraphics[width=5cm,height=5cm]{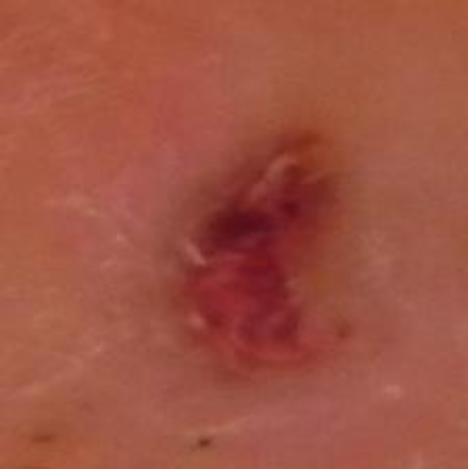} & \includegraphics[width=5cm,height=5cm]{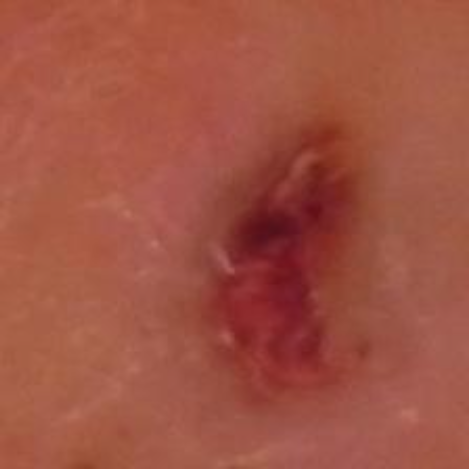} \\
		(a) & (b) \\
		\includegraphics[width=5cm,height=5cm]{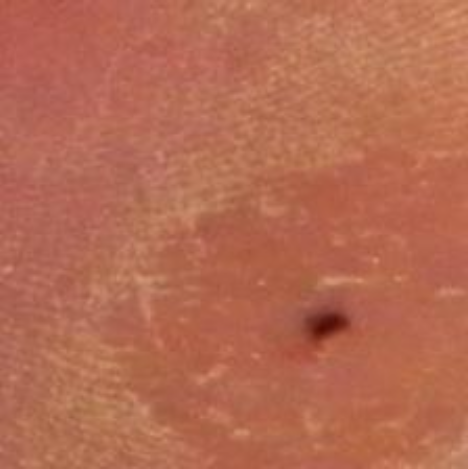} & \includegraphics[width=5cm,height=5cm]{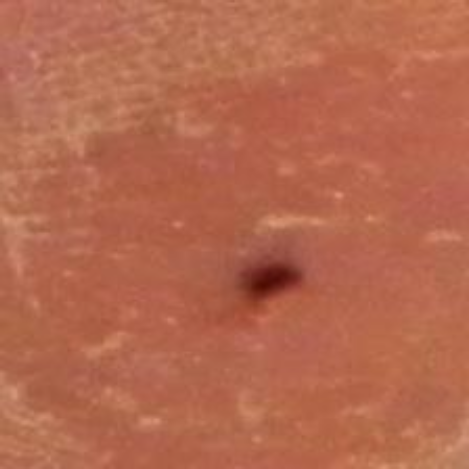} \\
		(c) & (d) \\
	\end{tabular}
	\caption[]{Illustration of two test set images identified by the dupeGuru fuzzy algorithm with a similarity threshold of 80\% (a \& b), and two test set images identified with a similarity threshold of 65\% (c \& d).}
	\label{fig:similar_test_set_65_and_80}
\end{figure}

Figure \ref{fig:similar_test_set_65_and_80} (a \& b) shows two similar images from the 80\% threshold. These images are of the same wound, with the second image being a natural augmentation case with a slightly different zoom level. The main visual differences can be observed on the bottom section of the image where the dark spots in image (a) are not present on image (b). Figure \ref{fig:similar_test_set_65_and_80} (c \& d) shows two similar images from the test set with 65\% similarity. As per the previous test examples (Figure \ref{fig:similar_test_set_65_and_80} (a \& b)), the second image represents a case of natural augmentation where the wound has a noticeably increased zoom level.

\subsection{Train \& Test Set Image Similarity}
\begin{figure}[]
	\centering
	\begin{tabular}{cc}
		\includegraphics[width=5cm,height=5cm]{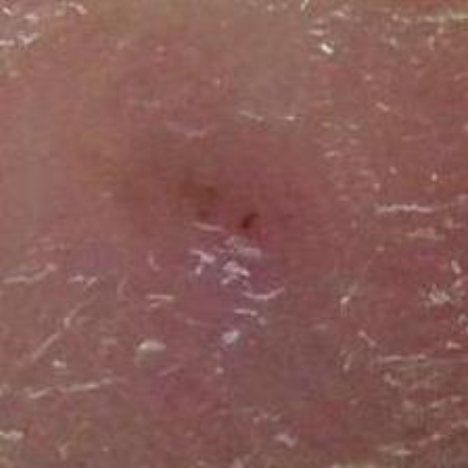} & \includegraphics[width=5cm,height=5cm]{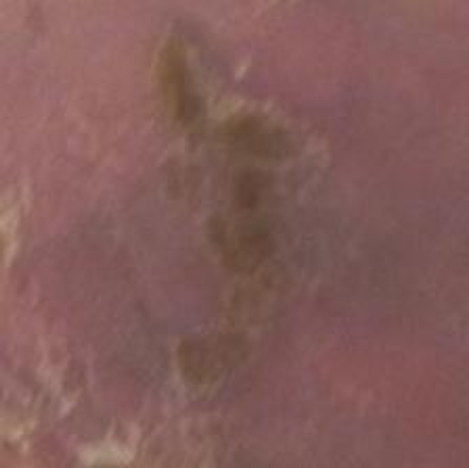} \\
		(a) & (b) \\
		\includegraphics[width=5cm,height=5cm]{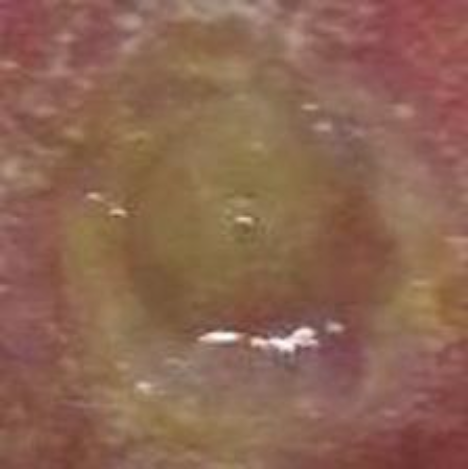} & \includegraphics[width=5cm,height=5cm]{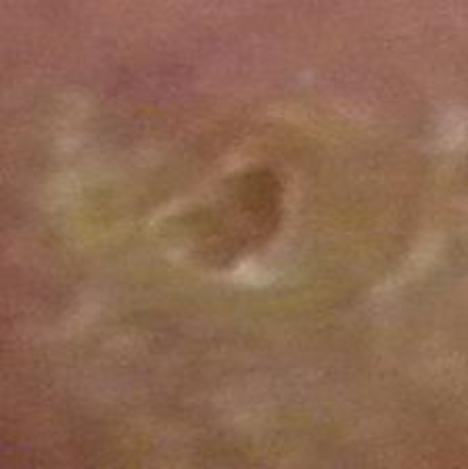} \\
		(c) & (d) \\
		\includegraphics[width=5cm,height=5cm]{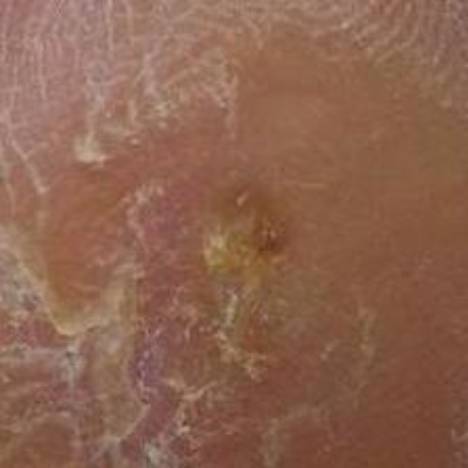} & \includegraphics[width=5cm,height=5cm]{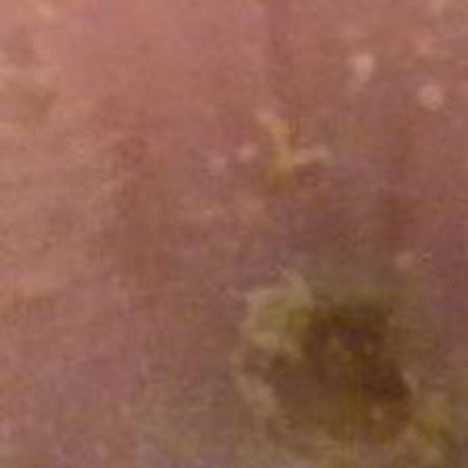} \\
		(e) & (f) \\
	\end{tabular}
	\caption[]{Illustration of similar images located across both training and test sets identified in the 60\% to 75\% similarity thresholds. Image (a) shows a training set image and image (b) shows a test set image, both identified in the 75\% similarity threshold.
		Image (c) shows a training set image and image (d) shows a test set image, both identified in the 65\% similarity threshold. Image (e) shows a training set image and image (f) shows a test set image, both identified in the 60\% similarity threshold.}
	\label{fig:similar_train_test_60_to_75}
\end{figure}

Figure \ref{fig:similar_train_test_60_to_75} (a \& b) shows two distinctly different wound images identified in the 75\% similarity threshold. Image (a) is from the training set, image (b) is from the test set. Figure \ref{fig:similar_train_test_60_to_75} (c \& d) and (e \& f) show further examples of visually similar images found across training and test sets at 65\%  and 60\% similarity thresholds respectively. Note that we do not discuss the class of the test set examples as these are part of a live public challenge for DFUC2021 which is still open to submissions (https://dfu-challenge.github.io/dfuc2021.html).

\subsection{Inter-class Image Similarity}
\begin{figure}[]
	\centering
	\begin{tabular}{cc}
		\includegraphics[width=5cm,height=5cm]{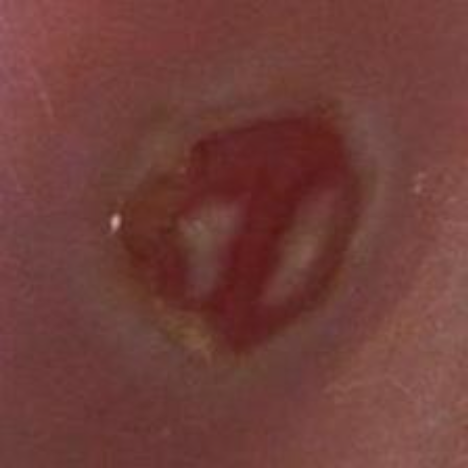} & \includegraphics[width=5cm,height=5cm]{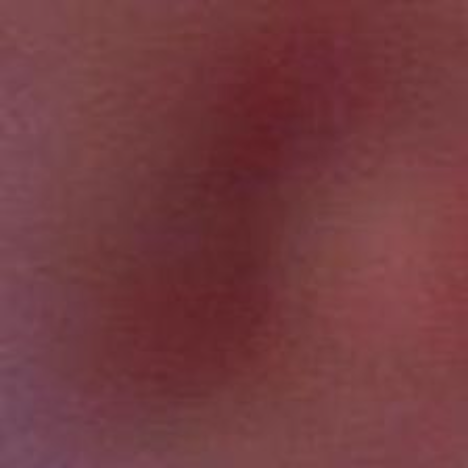} \\
		(a) & (b) \\
	\end{tabular}
	\caption[]{Illustration of two images from the training set which were identified in the inter-class similarity results for the 75\% similarity threshold: (a) an image from the `infection' class, and (b) an image from the `none' class}
	\label{fig:infection_none_nonidentical_75}
\end{figure}

\begin{figure}[]
	\centering
	\begin{tabular}{cc}
		\includegraphics[width=5cm,height=5cm]{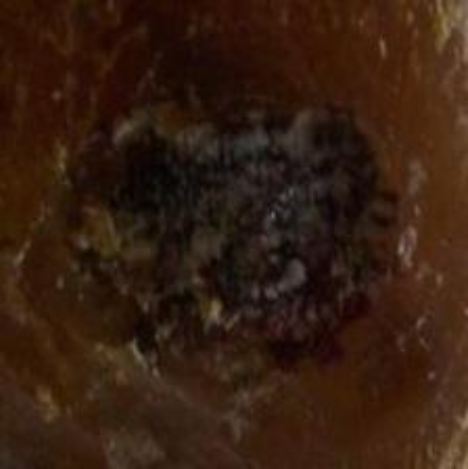} & \includegraphics[width=5cm,height=5cm]{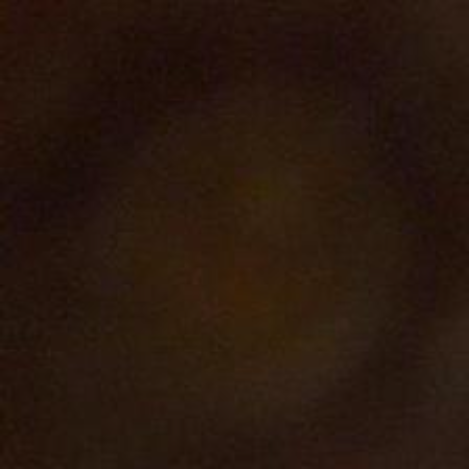} \\
	\end{tabular}
	\caption[]{Illustration of inter-class similarity results in the 70\% similarity threshold: (a) an image from the `ischemia' class, and (b) an image from the `none' class.}
	\label{fig:ischaemia_none_nonidentical_70}
\end{figure}

Our experiments using the dupeGuru fuzzy algorithm did not return any inter-class similarity results for the following groups of classes: (1) `both' vs `none', and (2) `infection' vs `ischemia'. For the `infection' vs `none' similarity searches, similar images were found for the 70\% (45 images), 75\% (13 images), and 80\% (4 images) similarity brackets. For the `ischemia' vs `none' similarity searches, similar images were found for the 70\% (5 images) and 75\% (3 images) brackets. Figure \ref{fig:infection_none_nonidentical_75} shows two images from the 75\% similarity threshold training set, with image (a) showing an `infection' class example, and image (b) showing a `none' class example. Figure \ref{fig:ischaemia_none_nonidentical_70} shows two images from the 70\% similarity threshold training set, where image (a) shows an `ischemia' class example, while image (b) shows a `none' class example.

\newpage

\subsection{Model Training}
Following the creation of each training set, as per Table \ref{tab:curated_sets}, we trained a selection of popular deep learning classification networks using each of our curated training sets. Batch size was set to 32 with stochastic gradient descent used for the optimiser and categorical cross-entropy as the loss function. Early stopping was implemented monitoring validation accuracy with a patience of 10. 
The hardware configuration used for our experiments was as follows: Intel Core i7-10750H CPU @2.60GHz, 64GB RAM, NVIDIA GeForce RTX 2070 Super with Max-Q Design 8GB. The software configuration used was as follows: Ubuntu 20.04 LTS, Python 3.8.13, and Tensorflow 2.4.2.

\section{Results and Discussion} \label{result}
This section details the results of training the multi-class classification networks on each of the training sets as detailed in Table \ref{tab:curated_sets}. We report the validation results from the full training set and the training sets using the following similarity thresholds: (1) 60\%, (2) 65\%, (3) 70\%, (4) 75\%, (5) 80\%, (6) 85\%, (7) 90\%, and (8) 95\%. Finally, we report the test results for the model with the highest validation accuracy.

\subsection{Baseline Results}
Table \ref{tab:full_set} shows the validation accuracy results for the models trained using the full training set, with no similar images removed. The InceptionResNetV2 model shows a clear lead in validation accuracy with 0.801, showing an increase of 0.038 over the next best performing model, which was ResNet50 with a validation accuracy of 0.763.

\begin{table}[htb]
	\caption{Best epoch and validation accuracy of the models trained on the full DFUC2021 dataset.}
	\centering
	\begin{tabular}{|l|l|l|}
		\hline
		\textbf{Model} &  \textbf{Best Epoch} & \textbf{Validation Accuracy} \\
		\hline 
		DenseNet201 & 36 & 0.731 \\
		\hline 
		EfficientNetB0 & 25 & 0.656 \\
		\hline
		EfficientNetB1 & 11 & 0.587 \\
		\hline
		EfficientNetB3 & 38 & 0.649 \\
		\hline
		\textbf{InceptionResNetV2} & 42 & \textbf{0.801} \\
		\hline
		InceptionV3 & 16 & 0.704 \\
		\hline
		\textbf{ResNet50} & 47 & \textbf{0.752} \\
		\hline
		ResNet50V2 & 43 & 0.738 \\
		\hline
		ResNet101 & 30 & 0.719 \\
		\hline
		ResNet101V2 & 54 & 0.735 \\
		\hline
		ResNet152 & 19 & 0.703 \\
		\hline
		\textbf{ResNet152V2} & 76 & \textbf{0.763} \\
		\hline
		VGG16 & 52 & 0.687 \\
		\hline
		VGG19 & 75 & 0.716 \\
		\hline
		Xception & 18 & 0.735 \\
		\hline
	\end{tabular}
	\label{tab:full_set}
\end{table}

\subsection{Results on the Curated Datasets}

The validation results for each of our curated training sets are shown in Table \ref{tab:60_set}, Table \ref{tab:65_set}, Table \ref{tab:70_set}, Table \ref{tab:75_set}, Table \ref{tab:80_set}, Table \ref{tab:85_set}, Table \ref{tab:90_set}, and Table \ref{tab:95_set}. The best performing model for validation accuracy is InceptionResNetV2 on the 80\% similarity threshold (0.885), as shown in Table \ref{tab:80_set}. This represents an increase of 0.030 over the next best performing model across all the best performing models in all similarity thresholds, which was InceptionResNetV2 in the 75\% similarity threshold with 0.855 accuracy as shown in Table \ref{tab:75_set}.

\begin{table}[]
	\caption{Best Epoch and validation accuracy of the models trained on the 60\% similarity threshold dataset.}
	\centering
	\begin{tabular}{|l|l|l|}
		\hline
		\textbf{Model} &  \textbf{Best Epoch} & \textbf{Validation Accuracy} \\
		\hline 
		DenseNet121 & 2 & 0.399 \\
		\hline
		DenseNet169 & 2 & 0.419 \\
		\hline 
		DenseNet201 & 3 & 0.403 \\
		\hline 
		EfficientNetB0 & 5 & 0.445 \\
		\hline
		EfficientNetB1 & 4 & 0.424 \\
		\hline
		EfficientNetB3 & 7 & 0.443 \\
		\hline
		\textbf{InceptionResNetV2} & 3 & \textbf{0.470} \\
		\hline
		\textbf{InceptionV3} & 3 & \textbf{0.461} \\
		\hline
		ResNet50 & 3 & 0.431 \\
		\hline
		ResNet50V2 & 8 & 0.416 \\
		\hline
		ResNet101 & 9 & 0.410 \\
		\hline
		ResNet101V2 & 1 & 0.421 \\
		\hline
		\textbf{ResNet152} & 2 & \textbf{0.460} \\
		\hline
		ResNet152V2 & 2 & 0.432 \\
		\hline
		VGG16 & 10 & 0.435 \\
		\hline
		VGG19 & 14 & 0.442 \\
		\hline
		Xception & 5 & 0.416 \\
		\hline
	\end{tabular}
	\label{tab:60_set}
\end{table}

\begin{table}[]
	\caption{Best Epoch and validation accuracy of the models trained on the 65\% similarity threshold dataset.}
	\centering
	\begin{tabular}{|l|l|l|}
		\hline
		\textbf{Model} &  \textbf{Best Epoch} & \textbf{Validation Accuracy} \\
		\hline 
		DenseNet121 & 19 & 0.673 \\
		\hline
		DenseNet169 & 33 & 0.731 \\
		\hline 
		DenseNet201 & 21 & 0.715 \\
		\hline 
		EfficientNetB0 & 44 & 0.670 \\
		\hline
		EfficientNetB1 & 53 & 0.664 \\
		\hline
		EfficientNetB3 & 28 & 0.662 \\
		\hline
		\textbf{InceptionResNetV2} & 54 & \textbf{0.823} \\
		\hline
		\textbf{InceptionV3} & 37 & \textbf{0.756} \\
		\hline
		\textbf{ResNet50} & 54 & \textbf{0.754} \\
		\hline
		ResNet50V2 & 35 & 0.742 \\
		\hline
		ResNet101 & 32 & 0.729 \\
		\hline
		ResNet101V2 & 17 & 0.671 \\
		\hline
		ResNet152 & 26 & 0.713 \\
		\hline
		ResNet152V2 & 55 & 0.744 \\
		\hline
		VGG16 & 21 & 0.651 \\
		\hline
		VGG19 & 42 & 0.662 \\
		\hline
		Xception & 26 & 0.733 \\
		\hline
	\end{tabular}
	\label{tab:65_set}
\end{table}

\begin{table}[]
	\caption{Best Epoch and validation accuracy of the models trained on the 70\% similarity threshold dataset.}
	\centering
	\begin{tabular}{|l|l|l|}
		\hline
		\textbf{Model} &  \textbf{Best Epoch} & \textbf{Validation Accuracy} \\
		\hline 
		\textbf{DenseNet121} & 60 & \textbf{0.765} \\
		\hline
		DenseNet169 & 34 & 0.707 \\
		\hline 
		DenseNet201 & 15 & 0.686 \\
		\hline 
		EfficientNetB0 & 30 & 0.693 \\
		\hline
		EfficientNetB1 & 27 & 0.660 \\
		\hline
		EfficientNetB3 & 7 & 0.581 \\
		\hline
		\textbf{InceptionResNetV2} & 29 & \textbf{0.759} \\
		\hline
		InceptionV3 & 26 & 0.715 \\
		\hline
		ResNet50 & 7 & 0.650 \\
		\hline
		ResNet50V2 & 18 & 0.684 \\
		\hline
		ResNet101 & 27 & 0.719 \\
		\hline
		ResNet101V2 & 28 & 0.684 \\
		\hline
		ResNet152 & 27 & 0.707 \\
		\hline
		ResNet152V2 & 14 & 0.660 \\
		\hline
		VGG16 & 28 & 0.669 \\
		\hline
		VGG19 & 34 & 0.686 \\
		\hline
		\textbf{Xception} & 48 & \textbf{0.791} \\
		\hline
	\end{tabular}
	\label{tab:70_set}
\end{table}

\begin{table}[]
	\caption{Best Epoch and validation accuracy of the models trained on the 75\% similarity threshold dataset.}
	\centering
	\begin{tabular}{|l|l|l|}
		\hline
		\textbf{Model} &  \textbf{Best Epoch} & \textbf{Validation Accuracy} \\
		\hline 
		DenseNet121 & 27 & 0.734 \\
		\hline
		DenseNet169 & 12 & 0.692 \\
		\hline 
		DenseNet201 & 16 & 0.684 \\
		\hline 
		EfficientNetB0 & 56 & 0.704 \\
		\hline
		EfficientNetB1 & 17 & 0.634 \\
		\hline
		EfficientNetB3 & 71 & 0.73 \\
		\hline
		\textbf{InceptionResNetV2} & 93 & \textbf{0.855} \\
		\hline
		InceptionV3 & 18 & 0.717 \\
		\hline
		ResNet50 & 24 & 0.706 \\
		\hline
		\textbf{ResNet50V2} & 45 & \textbf{0.747} \\
		\hline
		\textbf{ResNet101} & 32 & \textbf{0.74} \\
		\hline
		ResNet101V2 & 19 & 0.675 \\
		\hline
		ResNet152 & 28 & 0.706 \\
		\hline
		ResNet152V2 & 61 & 0.743 \\
		\hline
		VGG16 & 50 & 0.70 \\
		\hline
		VGG19 & 58 & 0.712 \\
		\hline
		Xception & 17 & 0.734 \\
		\hline
	\end{tabular}
	\label{tab:75_set}
\end{table}

\begin{table}[]
	\caption{Best epoch and validation accuracy of the models trained on the 80\% similarity threshold dataset.}
	\centering
	\begin{tabular}{|l|l|l|}
		\hline
		\textbf{Model} &  \textbf{Best Epoch} & \textbf{Validation Accuracy} \\
		\hline 
		DenseNet121 & 8 & 0.662 \\
		\hline
		DenseNet169 & 44 & 0.759 \\
		\hline 
		DenseNet201 & 20 & 0.704 \\
		\hline 
		EfficientNetB0 & 60 & 0.708 \\
		\hline
		EfficientNetB1 & 15 & 0.619 \\
		\hline
		EfficientNetB3 & 34 & 0.681 \\
		\hline
		\textbf{InceptionResNetV2} & 99 & \textbf{0.885} \\
		\hline
		\textbf{InceptionV3} & 43 & \textbf{0.769} \\
		\hline
		ResNet50 & 43 & 0.765 \\
		\hline
		ResNet50V2 & 32 & 0.722 \\
		\hline
		ResNet101 & 46 & 0.747 \\
		\hline
		\textbf{ResNet101V2} & 86 & \textbf{0.789} \\
		\hline
		ResNet152 & 45 & 0.733 \\
		\hline
		ResNet152V2 & 38 & 0.736 \\
		\hline
		VGG16 & 25 & 0.674 \\
		\hline
		VGG19 & 22 & 0.672 \\
		\hline
		Xception & 30 & 0.765 \\
		\hline
	\end{tabular}
	\label{tab:80_set}
\end{table}

\begin{table}[]
	\caption{Best epoch and validation accuracy of the models trained on the 85\% similarity threshold dataset.}
	\centering
	\begin{tabular}{|l|l|l|}
		\hline
		\textbf{Model} &  \textbf{Best Epoch} & \textbf{Validation Accuracy} \\
		\hline 
		DenseNet121 & 15 & 0.688 \\
		\hline
		DenseNet169 & 22 & 0.709 \\
		\hline 
		DenseNet201 & 27 & 0.742 \\
		\hline 
		EfficientNetB0 & 47 & 0.700 \\
		\hline
		EfficientNetB1 & 14 & 0.589 \\
		\hline
		EfficientNetB3 & 16 & 0.605 \\
		\hline
		\textbf{InceptionResNetV2} & 52 & \textbf{0.805} \\
		\hline
		InceptionV3 & 28 & 0.707 \\
		\hline
		ResNet50 & 53 & 0.748 \\
		\hline
		\textbf{ResNet50V2} & 64 & \textbf{0.767} \\
		\hline
		ResNet101 & 35 & 0.717 \\
		\hline
		ResNet101V2 & 50 & 0.735 \\
		\hline
		\textbf{ResNet152} & 61 & \textbf{0.755} \\
		\hline
		ResNet152V2 & 22 & 0.683 \\
		\hline
		VGG16 & 61 & 0.683 \\
		\hline
		VGG19 & 44 & 0.682 \\
		\hline
		Xception & 13 & 0.715 \\
		\hline
	\end{tabular}
	\label{tab:85_set}
\end{table}

\begin{table}[]
	\caption{Best epoch and validation accuracy of the models trained on the 90\% similarity threshold dataset.}
	\centering
	\begin{tabular}{|l|l|l|}
		\hline
		\textbf{Model} &  \textbf{Best Epoch} & \textbf{Validation Accuracy} \\
		\hline 
		DenseNet121 & 20 & 0.693 \\
		\hline
		DenseNet169 & 36 & 0.740 \\
		\hline 
		DenseNet201 & 14 & 0.700 \\
		\hline 
		EfficientNetB0 & 20 & 0.653 \\
		\hline
		EfficientNetB1 & 40 & 0.658 \\
		\hline
		EfficientNetB3 & 40 & 0.659 \\
		\hline
		\textbf{InceptionResNetV2} & 41 & \textbf{0.785} \\
		\hline
		\textbf{InceptionV3} & 57 & \textbf{0.768} \\
		\hline
		ResNet50 & 35 & 0.736 \\
		\hline
		ResNet50V2 & 23 & 0.699 \\
		\hline
		\textbf{ResNet101} & 46 & \textbf{0.743} \\
		\hline
		ResNet101V2 & 10 & 0.665 \\
		\hline
		ResNet152 & 28 & 0.715 \\
		\hline
		ResNet152V2 & 10 & 0.649 \\
		\hline
		VGG16 & 61 & 0.698 \\
		\hline
		VGG19 & 61 & 0.689 \\
		\hline
		Xception & 6 & 0.669 \\
		\hline
	\end{tabular}
	\label{tab:90_set}
\end{table}

\begin{table}[]
	\caption{Best epoch and validation accuracy of the models trained on the 95\% similarity threshold dataset.}
	\centering
	\begin{tabular}{|l|l|l|}
		\hline
		\textbf{Model} &  \textbf{Best Epoch} & \textbf{Validation Accuracy} \\
		\hline 
		\textbf{DenseNet121} & 78 & \textbf{0.817} \\
		\hline
		\textbf{DenseNet169} & 34 & \textbf{0.740} \\
		\hline 
		DenseNet201 & 24 & 0.720 \\
		\hline 
		EfficientNetB0 & 34 & 0.677 \\
		\hline
		EfficientNetB1 & 31 & 0.664 \\
		\hline
		EfficientNetB3 & 12 & 0.595 \\
		\hline
		InceptionResNetV2 & 21 & 0.734 \\
		\hline
		InceptionV3 & 28 & 0.726 \\
		\hline
		ResNet50 & 46 & 0.737 \\
		\hline
		ResNet50V2 & 39 & 0.726 \\
		\hline
		ResNet101 & 38 & 0.728 \\
		\hline
		ResNet101V2 & 51 & 0.737 \\
		\hline
		\textbf{ResNet152} & 107 & \textbf{0.820} \\
		\hline
		ResNet152V2 & 15 & 0.690 \\
		\hline
		VGG16 & 37 & 0.677 \\
		\hline
		VGG19 & 70 & 0.715 \\
		\hline
		Xception & 22 & 0.730 \\
		\hline
	\end{tabular}
	\label{tab:95_set}
\end{table}

The lowest performing models across all similarity thresholds for validation accuracy were ResNet152 (0.460), InceptionV3 (0.461), and InceptionResNetV2 (0.470), all present in the 60\% similarty threshold (see Table \ref{tab:60_set}). This indicates that the 60\% similarity threshold removed too many useful examples that the models were able to learn from - 2066 images compared to 317 images for the best performing network (InceptionResNetV2 at 80\% similarity threshold) in validation accuracy. All models trained in the 60\% similarity threshold also show low convergence for best epoch when compared to all other similarity thresholds, further highlighting a lack of learnable features present in this heavily curated training set.

Given that the InceptionResNetV2 model trained on the 80\% similarity threshold training set performed best in validation accuracy, we used this model to obtain test results on the DFUC2021 testing set. The results for this experiment are presented in Table \ref{tab:test_results}. The InceptionResNetV2 model trained on the 80\% similarity threshold training set shows clear performance improvements for macro average F1-score, precision, and recall, with improvements of 0.023, 0.029, 0.013 respectively. The reported AUC is slightly higher for the model trained on the full training set, however, this value is negligible with a difference of just 0.001.

\begin{table}[]
	\caption{Macro average test performance metrics for the InceptionResNetV2 model trained on the full DFUC2021 training set and the InceptionResNetV2 model trained on the 80\% similarity threshold training set. AUC - area under the curve.}
	\centering
	\begin{tabular}{|l|l|l|l|l|}
		\hline
		\textbf{Model} &  \textbf{F1-Score} & \textbf{Precision} & \textbf{Recall} & \textbf{AUC} \\
		\hline 
		InceptionResNetV2 (full) & 0.511 & 0.523 & 0.541 & \textbf{0.841} \\
		\hline 
		InceptionResNetV2 (80) & \textbf{0.534} & \textbf{0.552} & \textbf{0.554} & 0.840 \\
		\hline
	\end{tabular}
	\label{tab:test_results}
\end{table}

\begin{table}[]
	\caption{F1-score multi-class test results for the InceptionResNetV2 model trained on the full DFUC2021 training set and the InceptionResNetV2 model trained on the 80\% similarity threshold training set.}
	\centering
	\begin{tabular}{|l|l|l|l|l|l|}
		\hline
		\textbf{Model} &  \textbf{None} & \textbf{Infection} & \textbf{Ischemia} & \textbf{Both} & \textbf{Accuracy} \\
		\hline 
		Full & 0.707 & 0.512 & 0.431 & 0.394 & 0.602 \\
		\hline 
		80\% & \textbf{0.718} & \textbf{0.537} & \textbf{0.446} & \textbf{0.433} & \textbf{0.621} \\
		\hline
	\end{tabular}
	\label{tab:f1_scores}
\end{table}

The F1-scores for the multi-class test performance of the InceptionResNetV2 model trained on the 80\% similarity threshold training set are shown in Table \ref{tab:f1_scores}. The InceptionResNetV2 model trained on the 80\% similarity threshold training set shows a clear performance increase for all classes, with improvements of 0.011 for the none class, 0.025 for the infection class, 0.015 for the ischemia class, and 0.039 for the both class (infection and ischemia). The biggest performance increase is shown for the both class (0.039). Accuracy for all classes is 0.602 for the model trained on the full DFUC2021 training set, and 0.621 for the model trained on the 80\% similarity threshold training set. This demonstrates an accuracy improvement of 0.019 when testing using the InceptionResNetV2 model trained on the 80\% similarity threshold training set. 

We observe that a number of the visually similar images identified by the dupeGuru fuzzy algorithm were examples of natural augmentation. Our findings indicate that the excess use of subtle augmentation cases does not have the desired effect of boosting network performance. This highlights the importance of rigorously experimenting using individual augmentation sets when training deep learning networks to ascertain if models are being negatively affected by certain augmentation types. We encourage researchers working in other deep learning domains to follow these guidelines in future work to ensure that models are effectively trained, and that the effect of individual augmentation types is better understood.

Our experiments focused on the use of a single image similarity algorithm - an open-source fuzzy algorithm found in the dupeGuru application. Future research might test other image similarity methods, such as the structural similarity index measure, cosine similarity, or mean squared error \cite{cassidy2022isic}.

\section{Conclusion} \label{conclusion}
In this work we observed and quantified the effect of non-identical similar images on a selection of popular deep learning multi-class classification networks trained using a large publicly available diabetic foot ulcer dataset. We found that model accuracy is negatively affected by the presence of non-identical visually similar images, but that the removal of too many non-identical visually similar images can degrade network performance. We report our findings to encourage researchers to experiment with other deep learning datasets to gauge a better understanding of the effect of image similarity and the potential bias it may introduce into models trained on such data. 

\section*{Acknowledgment}
We gratefully acknowledge the support of NVIDIA Corporation who provided access to GPU resources for the DFUC2020 and DFUC2021 Challenges.


\bibliographystyle{unsrt}
\bibliography{references}

\end{document}